\documentclass[11pt]{article}

\PassOptionsToPackage{table}{xcolor}
\PassOptionsToPackage{hyperfootnotes=false}{hyperref}
\usepackage[final]{acl}
\usepackage{times}
\usepackage{latexsym}
\usepackage[T1]{fontenc}
\usepackage[utf8]{inputenc}
\usepackage{microtype}
\usepackage{inconsolata}
\usepackage{graphicx}
\usepackage{booktabs}
\usepackage{amsmath}
\usepackage{multirow}
\usepackage{enumitem}

\hypersetup{
  pdftitle={A Scaled-Up Empirical Study of Syntactic Language Models},
  pdfauthor={Pengcheng Wang, Zhexuan Zhang, Chao Lou, and Kewei Tu}
}

\title{A Scaled-Up Empirical Study of Syntactic Language Models}

\author{
  Pengcheng Wang\textsuperscript{1,2} \quad
  Zhexuan Zhang\textsuperscript{1,2} \quad
  Chao Lou\textsuperscript{3} \quad
  Kewei Tu\textsuperscript{1,2}\thanks{Corresponding author.}
  \\
  \textsuperscript{1}School of Information Science and Technology, ShanghaiTech University
  \\
  \textsuperscript{2}Shanghai Engineering Research Center of Intelligent Vision and Imaging
  \\
  \textsuperscript{3}Independent researcher
  \\
  \texttt{\{wangpch2024, zhangzhx2023, tukw\}@shanghaitech.edu.cn}
  \\
  \texttt{louchaobrn@outlook.com}
}

\begin{document}
\maketitle

\begin{abstract}
Syntactic language models (SLMs) that generate sentences along with their syntactic tree structures have shown promise in both syntactic generalization and downstream tasks. %%Some of these models enforce hierarchical attention patterns using structural attention masks. But two questions remain: whether the benefit comes from tree-structured input, structural attention masks, or both; and whether the approach scales to larger models and diverse tasks.
Prior studies, however, have mostly evaluated smaller models trained on less pretraining data and tested on narrower task suites. In addition, the design space of tree linearization and attention masks remains underexplored. 
We conduct a large-scale empirical study of 12 controlled configurations---one terminal baseline and 11 SLM variants spanning different linearizations and masks---at scales up to 1B parameters and 100B training tokens, evaluated across 11 downstream tasks and two syntactic benchmarks.
The experimental results show that SLMs with standard causal attention match or exceed structural-mask variants across downstream and syntactic benchmarks. They require no architectural modifications and remain compatible with FlashAttention. 
%The benefit comes from the input format, not the mask.
% We scale our method up with larger parameter sizes and diverse corpora, evaluating on more downstream tasks. 
% At the 1B scale, our method achieves the highest 11-task average.
We also observe that when scaling up to the 1B-parameter scale, the top-performing SLM variant no longer shows a clear advantage over a compute-matched non-structural baseline, suggesting that SLMs may become less competitive with scale. Our code is available at \url{https://github.com/qlwpc/TG-Interpolation}.
\end{abstract}

\section{Introduction}

Standard causal language models process text as flat token sequences. Although they can learn distributional proxies for syntax~\cite{hu-etal-2020-systematic}, hierarchical constituency structures---phrases composing into sentences---are left implicit. %%Providing explicit syntactic structure during pretraining could improve syntactic generalization and downstream reasoning. \cite{sartran-etal-2022-transformer}
Syntactic language models (SLMs)~\cite{sartran-etal-2022-transformer, hu-etal-2024-generative, murty-etal-2023-pushdown, zhao-etal-2024-dependency, huang2026GiLT} inject syntactic biases into Transformers~\cite{vaswani2017attention} using a simple technique: jointly modeling surface sentences and linearized versions of their syntactic parse trees.
These models typically employ \emph{structural} attention masks that allow a non-terminal to compose its child tokens once they are generated by focusing its attention on them.
\citet{zhao-etal-2025-systematic} conduct a systematic empirical comparison of various design choices for SLMs, but they do not fully explore the complete design space of tree linearization and attention masks.
Moreover, prior SLM studies have mostly evaluated models up to 300M parameters, trained on up to 15B tokens, and tested on only two or three tasks.

In this paper, we conduct a larger-scale empirical study of SLMs, evaluating selected models with up to 1B parameters and approximately 100B terminal training tokens on 11 downstream tasks and two syntactic benchmarks. We also investigate new designs of tree linearization and attention masks in a controlled design space of 12 configurations: one terminal baseline and 11 SLM variants.

Our findings are as follows: (i) Tree linearization with standard causal attention outperforms structural-mask variants on generation tasks and provides strong syntactic performance. It requires no architectural modifications and remains compatible with FlashAttention. (ii) At the 1B-parameter scale, the top-performing SLM variant has similar average downstream accuracy to a non-structural baseline that inserts learned dummy tokens between text tokens to provide approximately matched computation, while SLMs perform better on syntactic evaluation. Explicit structure thus remains useful, but its benefits must be assessed against the additional computation introduced by structural tokens.

Our contributions are:
\begin{itemize}
    \item We scale SLM evaluation to 1B parameters and approximately 100B terminal training tokens, across 11 downstream tasks and two syntactic benchmarks.
    \item We systematically examine a controlled family of 12 configurations (one terminal baseline and 11 SLM variants) with varying tree linearization and attention mask strategies.
    \item We include compute-matched comparisons between SLMs and non-structural baselines to assess the benefits of explicit structure alongside additional token positions.
\end{itemize}

\section{Related Work}

Prior work on syntactic language models (SLMs) has explored a broad range of ways to incorporate linguistic structure into neural sequence modeling. Transformer Grammars (TG)~\cite{sartran-etal-2022-transformer} encode constituency trees as a sequence of actions and enforce hierarchical attention patterns. A subsequent study extends the idea to dependency trees~\cite{zhao-etal-2024-dependency}. Building on this line of work, \citet{zhao-etal-2025-systematic} provide the most systematic comparison of SLM design choices including those of tree linearization and composition mechanisms. These studies show that syntactic inductive bias can improve language modeling and syntactic evaluation, but they typically couple modeling of linearized trees with specialized attention mechanisms.

Other studies incorporate syntactic inductive bias through alternative architectural mechanisms rather than explicit representations. These approaches include latent structure induction during pretraining~\cite{hu-etal-2024-generative}, stack-based recursive state tracking~\cite{murty-etal-2023-pushdown}, and dependency-graph-informed attention mechanisms~\cite{huang2026GiLT}.

\section{Variants of Syntactic Language Models}

\subsection{Syntactic Language Models}

Let $\mathbf{x} = (x_1, \ldots, x_N)$ be a sentence of $N$ terminal tokens, and let $\mathbf{y}$ be a labeled constituency parse tree for $\mathbf{x}$, where each node carries a phrase category (e.g., \texttt{S}, \texttt{NP}, or \texttt{VP}) and spans a contiguous subsequence of $\mathbf{x}$.
A syntactic language model predicts the sentence and its structural representation through a linearized token sequence $\mathbf{z} = (z_1, \ldots, z_T)$, factorized autoregressively as
\begin{equation}
p(\mathbf{z} := \textsc{Linearize}(\mathbf{x}, \mathbf{y})) = \prod_{t=1}^{T} p(z_t \mid z_{<t})
\label{eq:joint}
\end{equation}
where $\textsc{Linearize}$ maps the sentence--tree pair into a flat sequence. When the mapping is one-to-one, $p(\mathbf{z})$ is the joint probability $p(\mathbf{x},\mathbf{y})$; lossy linearizations can assign the same sequence to multiple trees. Prior SLMs vary in the \textbf{linearization} that serializes the tree $\mathbf{y}$ into $\mathbf{z}$, and the \textbf{architecture} that parameterizes the conditional probabilities $p(z_t \mid z_{<t})$. In Transformer-based SLMs, this architectural choice is largely realized through the \textbf{attention mechanism}, which governs how syntactic relationships among tokens in
$\mathbf{z}$ are processed.

\subsection{Tree Linearization}
\label{sec:tree_linearization}

\begin{table*}[!t]
\centering
\renewcommand{\arraystretch}{1.2}
\resizebox{0.94\textwidth}{!}{%
\setlength{\tabcolsep}{4pt}
\begin{tabular}{lcccp{9cm}}
\toprule

\textbf{Name} & \textbf{Left($\mathbf{v}$)} & \textbf{Right($\mathbf{v}$)} & \textbf{Post-process} & \textbf{Example} \\
\midrule
\rowcolor{gray!15} 
\multicolumn{5}{l}{\textbf{Linearization}} \\
$\textsc{Lin}^1$ & ONT($v$) & CNT($v$) & - & \texttt{(S Cats (VP chase mice VP) S)} \\
$\textsc{Lin}^2$ & ONT($v$) & 2$\times$CNT($v$) & - & \texttt{(S Cats (VP chase mice VP) VP) S) S)}  \\
$\textsc{Lin}^3$ & ONT($v$) & 3$\times$CNT($v$) & - & \texttt{(S Cats (VP chase mice VP) VP) VP) S) S) S)}  \\
$\textsc{Lin}^1_{\text{--ont}}$ & $\emptyset$ & CNT($v$) & - & \texttt{Cats chase mice VP) S)}  \\
$\textsc{Lin}^1_{\text{--merge}}$ & ONT($v$) & CNT($v$) & Merge CNTs & \texttt{(S Cats (VP chase mice )} \\\midrule
\rowcolor{gray!15} 
\multicolumn{5}{l}{\textbf{Non-structural}} \\
$\textsc{Lin}^0$ & $\emptyset$ & $\emptyset$ & - & \texttt{Cats chase mice} \\
$\textsc{Lin}^1_{\text{shuf}}$  &  ONT($v$) & CNT($v$) & Shuffle & \texttt{(S VP) Cats (VP chase S) mice} \\
Pause-1 & $\emptyset$ & $\emptyset$ & Insert 1 <PAUSE> & \texttt{Cats <PAUSE> chase <PAUSE> mice} \\
Pause-2 & $\emptyset$ & $\emptyset$ & Insert 2 <PAUSE> & \texttt{Cats <PAUSE> <PAUSE> chase <PAUSE> <PAUSE> mice} \\
\bottomrule
\end{tabular}%
}
\caption{Overview of linearizations and non-structural token sequences used in this paper. ``Post-process'' refers to operations applied after linearization. ONT and CNT denote opening (e.g., \texttt{(S} and \texttt{(VP}) and closing non-terminal tokens (e.g., \texttt{S)} and \texttt{VP)}), respectively. $v$ indicates that the emitted token carries a node label. ``Merged CNT'' signifies that consecutive CNTs (ignoring labels) are collapsed into a single \texttt{)} token. Finally, $\textsc{Lin}^1_{\text{shuf}}$ is generated by shuffling the $\textsc{Lin}^1$ token sequence while preserving the original terminal order.}
\label{tab:linearizations}
\end{table*}

The function $\textsc{Linearize}(\mathbf{x}, \mathbf{y})$ embeds sentence $\mathbf{x}$ and tree $\mathbf{y}$ into token sequence $\mathbf{z}$. We explore several variants of linearization, all of which share a simple recursive form:
\begin{align}
    \textsc{Lin}(v) &:= \textsc{Left}(v) \notag \\
    &\quad \circ [\, \textsc{Lin}(c) \text{ for } c ~\textsc{in} ~\textsc{children}(v) \,] \notag \\ 
    &\quad \circ \textsc{Right}(v) \\[1ex]
    \textsc{Lin}(t) &:= t
\end{align}
where $v\in \mathbf{y}$ is a node, $t\in\mathbf{x}$ is a terminal token, and $\circ$ denotes concatenation. $\textsc{Left}(v)$ and $\textsc{Right}(v)$ produce \emph{structural tokens} to enclose the linearized children. $\textsc{children}(v)$ returns the ordered children of $v$ in the tree. 

Table~\ref{tab:linearizations} summarizes all linearization variants that we study. $\textsc{Lin}^1$, $\textsc{Lin}^2$ and $\textsc{Lin}^3$ are three variants of linearization, differing only in the number of structural tokens emitted by $\textsc{Right}(v)$. Specifically, $\textsc{Lin}^1$ generates sequences resembling the widely used bracket notation for constituency trees, but with additionally labeled closing brackets. $\textsc{Lin}^2$ was proposed by \citet{sartran-etal-2022-transformer} to define their structural attention mask (referred to as STACK/COMPOSE in our paper). $\textsc{Lin}^3$ is a variant designed to test whether providing more closure tokens helps. We also include two additional variants, $\textsc{Lin}^1_{\text{--ont}}$ and $\textsc{Lin}^1_{\text{--merge}}$, to enable a more systematic comparison. %We select these variants to methodically probe the core design axes of syntax-aware linearization: structural token budget, opening tokens, and per-constituent closure. 
Other variations (e.g., different traversal orders) are not included because they would introduce confounds that distract from our central goal: disentangling tree linearization from the attention mask design. %characterizing emergent adaptive computation.

Different linearizations introduce varying numbers of structural tokens, inherently altering the total sequence length for training and inference (Table~\ref{tab:token-budgets}). %For instance, $\textsc{Lin}^1$ and $\textsc{Lin}^2$ produce $2.4\times$ and $3.2\times$ more tokens per raw document, respectively, compared to the baseline $\textsc{Lin}^0$. 
To assess the contribution of structural information alongside additional computation, we include three non-structural controls. First, $\textsc{Lin}^1_{\text{shuf}}$ shuffles the structural tokens while preserving terminal order. The other two, Pause-1 and Pause-2, insert one or two copies of a dedicated learned \texttt{PAUSE} token~\cite{goyal2024think} after each terminal, providing approximately comparable token budgets.

\begin{table}[htbp]
  \centering
  \small
  \setlength{\tabcolsep}{3pt}
  \begin{tabular}{lccc}
    \toprule
    Dataset & $\textsc{Lin}^0$ & $\textsc{Lin}^1$ & $\textsc{Lin}^2$ \\
    \midrule
    BBC News & 0.801B & 1.963B & 2.545B \\
    FineWeb-Edu & $\sim$98B & $\sim$233B & $\sim$301B \\
    \bottomrule
  \end{tabular}
  \caption{Corpus tokens by linearization, including special tokens. BBC corpus construction and training budgets are detailed in Appendices~\ref{sec:training} and~\ref{sec:historical-bbc}.}
  \label{tab:token-budgets}
\end{table}

\begin{table*}[!t]
  \centering
  \setlength{\tabcolsep}{6pt}
  \begin{tabular}{cllll}
    \toprule
    \# & Variant & Linearization & Attention  \\
    \midrule
    1 & Terminal & $\textsc{Lin}^0$ & Causal  \\
    2 & Tree & $\textsc{Lin}^1$ & Causal \\
    3 & TGTree & $\textsc{Lin}^2$ & Causal  \\
    4 & TG & $\textsc{Lin}^2$ & STACK/COMPOSE  \\
    5 & TGNomask & $\textsc{Lin}^2$ & Nomask  \\
    6 & TGNomask-Aug$^*$ & $\textsc{Lin}^2$ & Nomask  \\
    7 & Tree-NoONT & $\textsc{Lin}^1_{\text{--ont}}$ & Causal \\
    8 & Tree-Compress & $\textsc{Lin}^1_{\text{--merge}}$ & Causal \\
    9 & Tree-TripleCNT & $\textsc{Lin}^3$ & Causal  \\
    10 & Tree-Shuffle & $\textsc{Lin}^1_{\text{shuf}}$ & Causal   \\
    11 & TGNomask-Mix-TG & $\textsc{Lin}^2$ & Mixing heads (Nomask + STACK/COMPOSE)  \\
    12 & TGTree-Mix-TG & $\textsc{Lin}^2$ & Mixing heads (Causal   + STACK/COMPOSE)  \\
    \bottomrule
  \end{tabular}
  \caption{The 12 configurations in our controlled design space: one terminal baseline and 11 SLM variants. ONT = opening non-terminal, CNT = closing non-terminal. ``Hybrid'' denotes mixing-head models. $^*$TGNomask-Aug differs from TGNomask: CNT$_2$ remains visible to subsequent attention in Aug, while TGNomask prevents subsequent tokens from attending to it.}
  \label{tab:variants}
\end{table*}

\subsection{Structural Attention}
\label{sec:structural_attn}

A Transformer-based SLM can also incorporate structural information directly into the model itself via specially-designed attention mechanisms. We study three attention mechanisms as listed below. 

\paragraph{Causal Attention} The joint $p(\mathbf{x}, \mathbf{y})$ is modeled through standard autoregressive factorization over $\mathbf{z} = \textsc{Linearize}(\mathbf{x}, \mathbf{y})$ with \textit{no} modification to the attention mechanism.
Constituency structures are encoded entirely in the token sequence; the model learns to route information through Opening Non-Terminal (ONT) and Closing Non-Terminal (CNT) using standard causal attention.

\paragraph{Structural Attention}
Structural attention factorizes $p(\mathbf{x}, \mathbf{y})$ autoregressively over the linearized tree sequence $z = \textsc{Lin}^2(\mathbf{x}, \mathbf{y})$ and uses structure-dependent masks. In the original Transformer Grammar (TG) formulation~\cite{sartran-etal-2022-transformer}, each CNT is duplicated into two actions. CNT$_1$ performs \textsc{COMPOSE} attention over only the current constituent's children, producing a composed representation. CNT$_2$ performs \textsc{STACK} attention over only the representations on the active parse stack. TG then hides the child representations from later positions, forcing them to access the constituent through CNT$_1$. \citet{zhao-etal-2025-systematic} isolate this post-composition masking and show that removing it yields consistently better performance. We call the resulting variant \textbf{Nomask}: it retains the \textsc{COMPOSE} mask at CNT$_1$ but gives all other positions causal access to earlier terminals, ONTs, and CNT$_1$ representations. Thus, ``no mask'' refers specifically to removing TG's post-composition mask, not to unrestricted bidirectional attention. Our default Nomask still hides CNT$_2$ states from later positions; Nomask-Aug exposes them as well.

\paragraph{Hybrid Attention} The hybrid attention mechanism factorizes $p(\mathbf{x}, \mathbf{y})$ autoregressively over $z = \textsc{Lin}^2(\mathbf{x}, \mathbf{y})$ and replaces standard causal attention with \textbf{head-level mask mixing}: rather than applying a single mask type across all attention heads, different heads receive different masks. Hybrid attention enables the model to combine multiple structural attention patterns within a single Transformer layer.
Since enumerating all possible head-level mask assignments is computationally infeasible, we restrict ourselves to a straightforward setting: half of the attention heads receive one type of structural mask, and the remaining heads use another (either causal or structural). This configuration provides a tractable way to study hybrid attention.

Table~\ref{tab:variants} summarizes the 12 controlled configurations across the tree-linearization and attention-mask designs studied in our framework.

\section{Terminal Format Evaluation}
\label{sec:terminal-format-evaluation}

This section describes how we use SLMs to score candidate completions in multiple-choice (MC) tasks, given parses of the context and continuation.
We select the candidate with the highest score, applying task-specific normalization where appropriate.
Appendix~\ref{sec:olmes-eval} describes the input construction and scoring details.

SLM continuations interleave terminal content with structural ONT and CNT tokens, creating a choice between scoring the complete sequence and scoring only terminal tokens.
Let $\mathbf{p}$ denote the evaluation context and $\mathbf{c}_j$ the linearized continuation for candidate $j$.
Let $\mathcal{T}(\mathbf{c}_j)$ and $\mathcal{N}(\mathbf{c}_j)$ denote its terminal and non-terminal indices, respectively.
The full score includes both token types, whereas the terminal score includes only answer-content tokens:
\begin{align}
S_{\text{full}}(\mathbf{c}_j \mid \mathbf{p}) &= \sum_{t \in \mathcal{T} \cup \mathcal{N}} \log p(c_{j,t} \mid \mathbf{p}, \mathbf{c}_{j,<t}) \\
S_{\text{term}}(\mathbf{c}_j \mid \mathbf{p}) &= \sum_{t \in \mathcal{T}} \log p(c_{j,t} \mid \mathbf{p}, \mathbf{c}_{j,<t})
\label{eq:decom}
\end{align}

We use terminal scoring to compare models on answer content while retaining the supplied parses as context. This keeps structural-token likelihoods from contributing directly to the answer score. Appendix~\ref{sec:decom} compares MC predictions and accuracy with and without structural-token likelihoods in the answer score.

\section{Experiments}

\subsection{Setup}

We train models from scratch for one epoch using the same OLMo-based~\cite{groeneveld-etal-2024-olmo} decoder-only Transformer family. Learning rates and batch sizes follow the Step Law~\cite{li2025predictable} using each representation's actual training token budget. Causal models use FlashAttention-2~\cite{dao2023flashattention2}, while structural and hybrid models use specialized attention kernels. Appendix~\ref{sec:training} provides the architectures and training hyperparameters.

Table~\ref{tab:variants} defines the design space for our linearization and attention-mask ablations. To account for the varying token budgets induced by different linearizations in both training and inference, we compare SLMs with pause-token baselines~\cite{goyal2024think}: $\textsc{Pause-}K = (x_1, \underbrace{p, \ldots, p}_{K}, x_2, \underbrace{p, \ldots, p}_{K}, x_3, \ldots)$, where $p$ is a dedicated learned \texttt{PAUSE} token. Pause-1 doubles the terminal sequence length, compared with approximately $2.4\times$ for $\textsc{Lin}^1$; Pause-2 triples it, compared with approximately $3.2\times$ for $\textsc{Lin}^2$.

We also include Pushdown~\cite{murty-etal-2023-pushdown} as an architectural reference in the 100M experiment. It jointly trains a language model and an attachment parser, whose predictions are used to determine the stack depths that bias attention. Our reimplementation shares the OLMo backbone with the other models.

\subsection{Evaluation Methods}
\label{sec:evaluation-methods}

\paragraph{Document-Level Language Modeling.}
\label{sec:doc-ppl}
Computing sentence probabilities for an SLM requires marginalizing over syntactic representations. Following \citet{sartran-etal-2022-transformer} and \citet{zhao-etal-2025-systematic}, we obtain up to 300 valid proposals from a CRF constituency parser~\cite{kitaev-klein-2018-constituency}. Let $\mathcal{Z}'_i$ be the set of distinct token sequences obtained by linearizing these proposals for sentence $\mathbf{x}_i$. We approximate its probability by
\begin{align}
\widetilde p(\mathbf{x}_i \mid \mathbf{h}_i)
&= \sum_{\mathbf{z} \in \mathcal{Z}'_i}
p(\mathbf{z} \mid \mathbf{h}_i)
\leq p(\mathbf{x}_i \mid \mathbf{h}_i).
\label{eq:marginal}
\end{align}
Here, $\mathbf{h}_i$ is the preceding document context. Each term scores all tokens in the linearized sequence (Eq.~\ref{eq:joint}). Because $\mathcal{Z}'_i$ covers only a subset of the valid linearized sequences for $\mathbf{x}_i$, the truncated sum is a lower bound on the sentence probability conditioned on $\mathbf{h}_i$.

To avoid marginalizing over all preceding histories, we retain the highest-probability sequence for each completed sentence:
\begin{equation}
\hat{\mathbf{z}}_j=\arg\max_{\mathbf{z}\in\mathcal{Z}'_j}p(\mathbf{z}\mid\mathbf{h}_j).
\end{equation}
The selected sequences form the prefix $\mathbf{h}_i=\hat{\mathbf{z}}_1\circ\cdots\circ\hat{\mathbf{z}}_{i-1}$, and we marginalize only the current sentence. Fixing this prefix introduces a further approximation: the bound in Eq.~\ref{eq:marginal} holds for that history, but does not establish a bound on the full document marginal.

For a corpus of $D$ documents, let $\mathbf{X}_d=(\mathbf{x}_{d,1},\ldots,\mathbf{x}_{d,N_d})$ denote document $d$, where $\mathbf{x}_{d,i}$ is its $i$-th sentence.
Document-level perplexity is then
\begin{align}
\mathcal{L}_K
&= \sum_{d=1}^{D}\sum_{i=1}^{N_d}
\log \widetilde p(\mathbf{x}_{d,i}\mid\mathbf{h}_{d,i}), \nonumber\\
\text{PPL}_{\text{syn}} &= \exp(-\mathcal{L}_K/M).
\label{eq:doc-ppl}
\end{align}
Here, $M$ counts terminal and EOS tokens across the corpus. Terminal and pause-token models use their terminal-token likelihoods with the same denominator. Full evaluation details appear in Appendix~\ref{sec:ppl-protocol-details}.

\paragraph{Evaluations.} We measure downstream performance with task-specific fine-tuning on XSum summarization~\cite{narayan-etal-2018-dont}, reporting ROUGE-1/2/L~\cite{lin-2004-rouge}, and on BoolQ~\cite{clark-etal-2019-boolq} for boolean question answering. For each fixed BBC pretrained checkpoint, we run five independent fine-tuning seeds and report mean $\pm$ sample standard deviation for both tasks. For FineWeb-Edu models, we adopt the OLMES task formats and completion/cloze formulation~\cite{gu-etal-2025-olmes}. Our 11-task suite contains WinoGrande~\cite{sakaguchi2020winogrande}, HellaSwag~\cite{zellers-etal-2019-hellaswag}, BoolQ, MMLU~\cite{hendrycks2021measuring}, MMLU-Redux~\cite{gema-etal-2025-done}, OpenBookQA~\cite{mihaylov-etal-2018-suit}, CommonsenseQA~\cite{talmor-etal-2019-commonsenseqa}, SocialIQA~\cite{sap-etal-2019-social}, ARC-Easy and ARC-Challenge~\cite{clark2018think}, and PIQA~\cite{bisk2020piqa}. We additionally evaluate syntactic competence on SG~\cite{hu-etal-2020-systematic} and BLiMP~\cite{warstadt-etal-2020-blimp-benchmark}.

\begin{table*}[!t]
  \centering
  \normalsize
  \setlength{\tabcolsep}{7pt}
  \begin{tabular}{lccccc}
    \toprule
    Model & XSum$\uparrow$ & BoolQ$\uparrow$ & SG$\uparrow$ & BLiMP$\uparrow$ & Doc-PPL$\downarrow$ \\
    \midrule
    \multicolumn{6}{l}{\textbf{BBC 100M}} \\
    \midrule
    Terminal-100M & 22.02 $\pm$ 0.06 & 66.70 $\pm$ 0.36 & 69.80 & 70.77 & 15.369 \\
    Tree-100M & 22.84 $\pm$ 0.03 & 68.07 $\pm$ 0.35 & 81.40 & 80.95 & 14.736 \\
    TGTree-100M & \textbf{23.03 $\pm$ 0.02} & 68.12 $\pm$ 0.62 & 79.47 & 81.46 & 14.709 \\
    TG-100M & 20.15 $\pm$ 0.04 & 64.50 $\pm$ 0.40 & 79.19 & 83.28 & 16.405 \\
    TGNomask-100M & 21.91 $\pm$ 0.04 & 66.66 $\pm$ 0.45 & 78.62 & 81.78 & 14.658 \\
    TGNomask-Aug-100M & 22.12 $\pm$ 0.02 & 67.01 $\pm$ 0.28 & 78.68 & \textbf{83.63} & 14.700 \\
    Tree-NoONT-100M & 22.41 $\pm$ 0.04 & \textbf{68.67 $\pm$ 0.32} & \textbf{85.38} & 81.00$^\ddagger$ & 15.079 \\
    Tree-Compress-100M & 22.70 $\pm$ 0.05 & 68.48 $\pm$ 0.49 & 78.27 & 81.40$^\ddagger$ & 14.748 \\
    Tree-TripleCNT-100M & 22.43 $\pm$ 0.04 & 68.09 $\pm$ 0.12 & 81.57 & 83.42 & 14.801 \\
    Tree-Shuffle-100M$^\dagger$ & 21.56 $\pm$ 0.09 & 68.08 $\pm$ 0.33 & 76.56 & 67.77 & 17.852 \\
    TGNomask-Mix-TG-100M & 21.87 $\pm$ 0.04 & 66.13 $\pm$ 0.54 & 75.77 & 83.14 & 14.573 \\
    TGTree-Mix-TG-100M & 22.16 $\pm$ 0.07 & 67.91 $\pm$ 0.36 & 78.84 & 82.75 & \textbf{14.482} \\
    Pause-1-100M & 22.38 $\pm$ 0.05 & 62.04 $\pm$ 0.19 & 75.11 & 70.85 & 14.916 \\
    Pause-2-100M & 22.25 $\pm$ 0.06 & 65.70 $\pm$ 3.45 & 76.97 & 72.59 & 15.116 \\
    \rowcolor{gray!10}
    Pushdown-100M & 21.06 $\pm$ 0.05 & 65.54 $\pm$ 0.29 & 74.86 & 76.76 & 19.316 \\
    \midrule
    \multicolumn{6}{l}{\textbf{BBC 500M}} \\
    \midrule
    Terminal-500M & 23.11 $\pm$ 0.07 & \textbf{70.54 $\pm$ 0.51} & 80.88 & 72.31 & 14.029 \\
    Tree-500M & 23.57 $\pm$ 0.06 & 70.05 $\pm$ 0.58 & 83.43 & 82.76 & 13.864 \\
    TGTree-500M & \textbf{23.59 $\pm$ 0.05} & 68.43 $\pm$ 0.26 & 80.68 & \textbf{83.72} & 13.947 \\
    Pause-1-500M & 23.42 $\pm$ 0.06 & 70.26 $\pm$ 0.41 & 79.00 & 70.85 & 14.019 \\
    Pause-2-500M & 23.38 $\pm$ 0.02 & 68.86 $\pm$ 0.32 & 78.18 & 72.69 & 14.030 \\
    Tree-NoONT-500M & 23.30 $\pm$ 0.03 & 70.24 $\pm$ 0.26 & \textbf{83.67} & 81.10$^\ddagger$ & 13.955 \\
    TG-500M & 20.67 $\pm$ 0.07 & 65.55 $\pm$ 0.30 & 79.55 & 83.32 & 15.465 \\
    TGNomask-500M & 22.14 $\pm$ 0.07 & 69.54 $\pm$ 0.61 & 81.58 & 83.21 & 13.995 \\
    TGTree-Mix-TG-500M & 23.31 $\pm$ 0.04 & 69.41 $\pm$ 0.33 & 79.76 & 83.37 & \textbf{13.785} \\
    TGNomask-Mix-TG-500M & 22.37 $\pm$ 0.06 & 69.06 $\pm$ 0.29 & 80.11 & 83.22 & 13.973 \\
    \bottomrule
  \end{tabular}
  \caption{BBC results at 100M and 500M. XSum reports R-AVG; XSum and BoolQ show mean $\pm$ sample SD over five fine-tuning seeds. Other metrics are single evaluations. Best values at each scale are in \textbf{bold}. The shaded Pushdown row is an architectural reference. $\dagger$~Tree-Shuffle uses the same PPL evaluation method as Terminal. $\ddagger$~marks BLiMP scores pending candidate-uniqueness checks (Appendix~\ref{sec:ppl-protocol-details}).}
  \label{tab:exp1-results}
\end{table*}

\subsection{Experiment 1: Disentangling Tree Linearization and Attention}
\label{sec:exp1}

For training corpus construction, we use the Benepar parser~\cite{kitaev-klein-2018-constituency}, which is trained on news-domain data, to obtain constituency parses for BBC News articles filtered from each subset of the FineWeb dataset~\cite{penedo2024fineweb}. The reference corpus contains 1.47M articles and 0.801B terminal tokens (Table~\ref{tab:token-budgets}). At 100M, we evaluate all 12 controlled configurations, two pause-token controls, and Pushdown; at 500M, we evaluate eight configurations and two pause-token controls. Table~\ref{tab:exp1-results} reports both experiments. Training-data details, the data audit, and supplementary BBC experiments appear in Appendices~\ref{sec:training} and~\ref{sec:historical-bbc}.

\paragraph{Structural Attention vs.\ Causal Attention.}
Tree and TGTree achieve the highest XSum means in both BBC settings. At 100M, they exceed TGNomask-Aug and TG, whereas TGNomask-Aug has the highest BLiMP score. At 500M, Tree and TGTree exceed TGNomask by more than 1.4 XSum points and TG by more than 2.8 points; TGTree also leads on BLiMP. TGNomask improves on TG in XSum, BoolQ, and perplexity in both settings, consistent with \citet{zhao-etal-2025-systematic}. Causal attention does not lead on every task: TGNomask exceeds TGTree on SG and BoolQ at 500M, and Terminal has the highest BoolQ mean in that setting.

\paragraph{Hybrid Attention.}
Mixing causal and structural heads improves TGTree's document perplexity from 14.709 to 14.482 at 100M and from 13.947 to 13.785 at 500M, giving the lowest value in each setting. At 500M, it also improves BoolQ by 0.98 points, but lowers XSum, SG, and BLiMP. Mixing TGNomask and TG heads likewise lowers perplexity in both settings. Hybrid attention is therefore useful for language modeling, even though it does not improve on causal Tree models for summarization.

\paragraph{Pause-token Controls.}
Pause tokens improve XSum over Terminal in both settings, indicating that additional token positions can help generation. Tree and TGTree nevertheless exceed their corresponding pause controls on XSum, SG, BLiMP, and document perplexity. At 500M, both tree models gain more than 11 BLiMP points. BoolQ favors the tree models at 100M but the pause controls at 500M, by 0.21 and 0.43 points. Thus, explicit structure offers benefits beyond those observed with pause tokens, particularly on syntactic evaluation.

\paragraph{Document-Level Perplexity.}
Tree and TGTree improve on Terminal at 100M (14.736 and 14.709 vs.\ 15.369) and 500M (13.864 and 13.947 vs.\ 14.029). The full STACK/COMPOSE mask performs worse, while TGNomask recovers much of the loss. NoONT obtains 15.079 at 100M and 13.955 at 500M; Compress obtains 14.748 at 100M. Both linearizations have higher perplexity than Tree in their respective settings. The lowest perplexity comes from hybrid attention, whereas the highest XSum mean comes from TGTree.

\begin{table*}[!t]
  \centering
  \setlength{\tabcolsep}{1.5pt}
  \begin{tabular}{lccccccccccccc}
    \toprule
    Model & WG$\uparrow$ & HS$\uparrow$ & BQ$\uparrow$ & MM$\uparrow$ & MMR$\uparrow$ & OBQA$\uparrow$ & CSQA$\uparrow$ & SIQA$\uparrow$ & ARCE$\uparrow$ & ARCC$\uparrow$ & PIQA$\uparrow$ & \textbf{AVG}$\uparrow$ \\
    \midrule
    Terminal & 58.17 & 61.53 & 58.65 & 36.79 & 37.61 & 43.60 & 53.24 & 48.98 & 69.30 & 43.48 & 73.88 & 53.20 \\
    Tree & 61.64 & 54.10 & 63.85 & 36.88 & 37.53 & 44.00 & 56.02 & 48.46 & 68.42 & 45.15 & 73.07 & 53.56 \\
    TGTree & \textbf{63.38} & 63.51 & \textbf{67.55} & 37.76 & 38.42 & \textbf{45.80} & \textbf{58.15} & 50.31 & 70.00 & 44.82 & 74.97 & \textbf{55.88} \\
    Pause-1 & 58.96 & 63.05 & 65.41 & 37.70 & 38.81 & 42.80 & 56.18 & \textbf{51.28} & \textbf{74.04} & 43.48 & \textbf{75.14} & 55.17 \\
    Pause-2 & 62.59 & \textbf{64.45} & 63.39 & \textbf{38.38} & \textbf{39.46} & 44.00 & 56.76 & 48.67 & 72.98 & \textbf{48.16} & 74.97 & 55.80 \\
    \bottomrule
  \end{tabular}
  \caption{Downstream accuracy at 1B on FineWeb-Edu using the terminal score $S_{\text{term}}$ defined in Section~\ref{sec:terminal-format-evaluation}. WG = WinoGrande, HS = HellaSwag, BQ = BoolQ, MM = MMLU, MMR = MMLU-Redux, OBQA = OpenBookQA, CSQA = CommonsenseQA, SIQA = SocialIQA, ARCE = ARC-Easy, ARCC = ARC-Challenge. Best per task in \textbf{bold}; Table~\ref{tab:bootstrap-ci} reports 95\% bootstrap intervals.}
  \label{tab:main-result}
\end{table*}

\paragraph{Tree Linearization Ablations.}
Removing opening non-terminals improves SG but lowers XSum in both settings, suggesting that opening tokens are more useful for generation than for the distinctions tested by SG. Doubling closing tokens gives the highest XSum mean in both experiments. At 100M, tripling them further improves SG and BLiMP over TGTree but lowers XSum; merging consecutive closing tokens lowers SG. Shuffling structural tokens causes a substantial BLiMP decline. The SG subtask results at 500M appear in Appendix~\ref{sec:sg-breakdown}.

\paragraph{Scaling Up.}
We study Tree and TGTree at 1B because of their generation performance and compatibility with efficient causal attention. They use the same attention architecture as Terminal and the pause-token baselines, enabling a direct comparison of input representations. Appendix~\ref{sec:historical-efficiency} reports throughput measurements for the earlier implementations.

\subsection{Experiment 2: Scaling to 1B on FineWeb-Edu}
\label{sec:exp2}

FineWeb-Edu-100BT~\cite{penedo2024fineweb} is a $\sim$100B-token sampled subset of the FineWeb-Edu corpus, filtered for educational quality and spanning diverse domains.
We train five models at 1B parameters on this data: Terminal, Tree, TGTree, Pause-1 and Pause-2.
For downstream evaluation, we use the 11-task OLMES suite~\cite{gu-etal-2025-olmes} with the terminal score $S_{\text{term}}$ defined in Section~\ref{sec:terminal-format-evaluation}. We also evaluate syntactic performance on SG and BLiMP. Table~\ref{tab:main-result} reports per-task downstream accuracy, and Table~\ref{tab:1b-syntax} reports the syntactic evaluation results. We quantify evaluation-set uncertainty using 10,000 paired bootstrap resamples; full confidence intervals and tests appear in Appendix~\ref{sec:bootstrap}, and Appendix~\ref{sec:decom} reports sensitivity to full versus terminal scoring.

\paragraph{Downstream Performance.}
TGTree achieves the highest average accuracy across the 11 tasks (55.88), exceeding Terminal by 2.68 points. Its gains over Terminal are significant on five tasks under paired bootstrap tests. Tree improves the average by only 0.36 points and performs particularly poorly on HellaSwag, suggesting that the choice of linearization remains important at this scale.

\paragraph{Explicit Structure vs.\  Dummy Tokens.}
TGTree and Pause-2 have nearly identical average accuracy (55.88 vs.\ 55.80), with complementary strengths. TGTree is significantly better on BoolQ, while Pause-2 is significantly better on HellaSwag; differences on the other nine tasks are not significant at $\alpha=.05$. Excluding BoolQ puts Pause-2 ahead (55.04 vs.\ 54.71). These results show that a strong non-structural baseline can achieve similar aggregate downstream accuracy to an SLM under approximately comparable token budgets. Appendix~\ref{sec:bootstrap} reports the paired comparisons, and Appendix~\ref{sec:validation10} examines an alternative task suite.

\begin{table}[tbp]
  \centering
  \setlength{\tabcolsep}{8pt}
  \begin{tabular}{lcc}
    \toprule
    Model & BLiMP$\uparrow$ & SG$\uparrow$ \\
    \midrule
    Terminal & 79.65 & 88.01 \\
    Tree & \textbf{81.79} & \textbf{89.95} \\
    TGTree & 80.98 & 89.00 \\
    Pause-1 & 80.83 & 86.54 \\
    Pause-2 & 80.36 & 86.59 \\
    \bottomrule
  \end{tabular}
  \caption{Syntactic evaluation at 1B on FineWeb-Edu. Evaluation details appear in Appendix~\ref{sec:eval-details}.}
  \label{tab:1b-syntax}
\end{table}

\paragraph{Syntactic Evaluation.}
Tree achieves the highest scores on both syntactic benchmarks, exceeding Terminal by 2.14 points on BLiMP and 1.94 points on SG (Table~\ref{tab:1b-syntax}). TGTree also improves both scores. Pause tokens improve BLiMP but lower SG relative to Terminal, leaving a larger gap to the SLMs on this benchmark. Although TGTree and Pause-2 have similar downstream averages, TGTree scores 0.62 points higher on BLiMP and 2.41 points higher on SG. The pause-token baseline therefore matches TGTree's aggregate downstream performance but not its syntactic performance.

\section{Discussion}

\subsection{Why Does Tree Linearization Help Without Structural Attention?}
\label{sec:why-works}

We propose three complementary hypotheses:

\paragraph{Implicit multi-task learning hypothesis.} The model jointly predicts terminal tokens and syntactic structure, making constituency boundaries explicit through non-terminals instead of inferring them only from co-occurrence statistics. \citet{sartran-etal-2022-transformer} attribute the perplexity gains of such joint modeling to the ``additional syntactic supervision'' that non-terminals provide by constraining admissible word classes at each position.

\paragraph{Compositional bottleneck hypothesis.} CNT positions may act as ``summary points'' that encourage compression of constituent information. Under this hypothesis, causal attention allows the model to learn which information to route through CNT positions, unlike TG's hard-coded COMPOSE attention. It can also preserve direct access to earlier words when a task requires lexical detail. We do not directly test this mechanism.

\paragraph{Adaptive computation hypothesis.} Pause-$K$ allocates a fixed number of learned latent tokens after each terminal. Tree models instead use parse-dependent structural tokens: richer constituency structures introduce more ONT/CNT positions at phrase boundaries. This placement may allocate additional computation near constituent composition. The gains of both tree and pause-token models over Terminal motivate studying how the number and placement of these positions affect learning.

\subsection{Syntactic Generalization and Downstream Performance}
\label{sec:sg-discussion}

The best linearization differs between syntactic evaluation and downstream tasks. In both BBC settings, Tree-NoONT has the highest SG score, whereas TGTree has the highest XSum mean. On FineWeb-Edu, Tree leads on both syntactic benchmarks, but TGTree has the higher downstream average. Predicting additional structural tokens therefore does not improve all forms of syntactic generalization equally.

The pause-token comparisons further distinguish these outcomes. Pause-2 approaches TGTree's average downstream accuracy at 1B, yet remains below it on SG and BLiMP. Additional token positions can support downstream performance without producing the same syntactic behavior as explicit tree tokens. This distinction motivates evaluating SLMs on both targeted syntactic tests and a broad downstream suite.

\section{Conclusion}

We conduct a scaled-up empirical study of syntactic language models, examining tree linearization and attention design and evaluating selected models up to 1B parameters. At both 100M and 500M on BBC News, causal tree models achieve strong generation and syntactic performance while remaining compatible with FlashAttention; hybrid attention provides the best document perplexity. At 1B on FineWeb-Edu, TGTree and Pause-2 have similar aggregate downstream performance, while tree models retain an advantage on syntactic evaluation. These findings support tree linearization as a practical way to introduce syntactic bias and highlight the need to compare its benefits with those of additional non-structural computation.

\section*{Limitations}

Constructing linearized tree inputs requires a parser for training and evaluation. This adds preprocessing cost, and parsing errors may propagate into the model. We do not quantify the effect of parse quality and study only English. The compositional bottleneck and adaptive computation accounts remain hypotheses; testing them would require interventions on the models' internal representations or attention patterns.

Pause-token controls approximately match sequence lengths rather than exact FLOPs. We do not evaluate structural-mask models at 1B on FineWeb-Edu.

\section*{Acknowledgments}
This work was supported by the HPC Platform of ShanghaiTech University, the SIST Computing Platform and the Core Facility Platform of Computer Science and Communication, SIST, ShanghaiTech University. 

\bibliography{anthology-1,anthology-2,custom}

\appendix

\section{Training Details}
\label{sec:training}

\subsection{Implementation and Hyperparameters}

We use an OLMo-based~\cite{groeneveld-etal-2024-olmo} decoder-only Transformer with SwiGLU~\cite{shazeer2020gluSWIGLU} activation, RoPE~\cite{su2024roformerROPE} positional encoding, RMSNorm~\cite{zhang2019rootRMS} for pre-normalization, and tied input and output embeddings.
Table~\ref{tab:scales} lists the three scale configurations. Scale names refer to non-embedding parameters; the 500M models contain 508M non-embedding parameters and 579M total parameters with tied embeddings.

\begin{table}[htbp]
  \centering
  \small
  \setlength{\tabcolsep}{3pt}
  \begin{tabular}{lccccc}
    \toprule
    Scale & $d_\text{model}$ & MLP ratio & Layers & Heads & Params \\
    \midrule
    100M & 768 & 4 & 12 & 12 & $\sim$100M \\
    500M & 1408 & 8 & 16 & 16 & $\sim$500M \\
    1B & 2048 & 8 & 16 & 16 & $\sim$1B \\
    \bottomrule
  \end{tabular}
  \caption{Model configurations at each scale.}
  \label{tab:scales}
\end{table}

We train all models from scratch for one epoch. Training uses the AdamW optimizer~\cite{loshchilov2017decoupledADAM} with $\beta_1{=}0.9$, $\beta_2{=}0.95$, $\epsilon{=}10^{-8}$, weight decay of 0.1, and gradient clipping at norm of 1.0. We use cosine learning-rate decay with 2,000 warmup steps and a minimum learning rate of $10^{-5}$. The maximum sequence length is 2,048 tokens except for BBC Pause-2 at 500M, whose 683 terminal positions expand to 2,049 positions. The learning rate $\eta$ and batch size $B$ are set according to the Step Law~\cite{li2025predictable}:

\begin{equation}
    \begin{gathered}
       \eta(N,D) = 1.79N^{-0.713}D^{0.307} \\
B(D) = 0.58D^{0.571}
    \end{gathered}
\label{eq:steplaw}
\end{equation}
where $N$ is the number of non-embedding parameters and $D$ is the actual training token budget, including repeated articles and structural or pause positions. This budget is distinct from the deduplicated BBC corpus size in Table~\ref{tab:token-budgets}. We convert $B(D)$ from tokens to sequences and round for distributed training. Table~\ref{tab:dedup-training} gives the BBC 500M settings. All ten runs use seed 6198, BF16, and two DDP replicas, with a per-device microbatch of four and gradient accumulation to the listed global batch. Training processes all complete batches for one epoch. Tree uses 1.226 times as many input positions as Pause-1, and TGTree uses 1.059 times as many as Pause-2; their learning rates and batch sizes also differ according to Eq.~\ref{eq:steplaw}.

\begin{table*}[!t]
  \centering
  \small
  \setlength{\tabcolsep}{8pt}
  \begin{tabular}{lrrrrr}
    \toprule
    Model & Training tokens $D$ & Peak LR ($\times10^{-4}$) & Global batch & Length & Final step \\
    \midrule
    Terminal & 800,848,405 & 6.012086 & 34 & 2048 & 11,501 \\
    Tree & 1,963,388,153 & 7.917501 & 58 & 2048 & 16,529 \\
    TGTree & 2,544,658,027 & 8.573605 & 66 & 2048 & 18,825 \\
    Pause-1 & 1,601,696,810 & 7.437747 & 52 & 2048 & 15,039 \\
    Pause-2 & 2,402,545,215 & 8.423672 & 64 & 2049 & 18,321 \\
    Tree-NoONT & 1,382,118,279 & 7.108576 & 46 & 2048 & 14,670 \\
    TG & 2,544,658,027 & 8.573605 & 66 & 2048 & 18,825 \\
    TGNomask & 2,544,658,027 & 8.573605 & 66 & 2048 & 18,825 \\
    TGTree-Mix-TG & 2,544,658,027 & 8.573605 & 66 & 2048 & 18,825 \\
    TGNomask-Mix-TG & 2,544,658,027 & 8.573605 & 66 & 2048 & 18,825 \\
    \bottomrule
  \end{tabular}
  \caption{Training settings for the ten BBC 500M models. Batch counts sequences, length counts input positions, and steps identify the final checkpoints used for evaluation. Pause budgets include inserted pause tokens.}
  \label{tab:dedup-training}
\end{table*}

\paragraph{Data Preparation.}
The deduplicated BBC reference corpus contains 1,466,132 articles from 89 crawl shards spanning 2013--2022. We remove exact terminal-sequence, normalized-text, and near-duplicate articles from the historical parsed crawl pool, then encode retained articles with PTB mapping and explicit BOS/EOS tokens. The terminal representation contains 800,848,405 tokens (0.801B); tree representations add the structural tokens listed in Table~\ref{tab:dedup-training}. We evaluate the final checkpoint from each run. Training loss includes structural or pause targets, whereas document PPL is normalized by terminal and EOS tokens. Data manifests, tokenizer definitions, and checkpoint identifiers are provided with the code.

\subsection{Computational Costs}
\label{sec:historical-efficiency}
The BBC 500M models in Table~\ref{tab:exp1-results} each train on two RTX 5090 GPUs. Causal models use FlashAttention-2, and TG, TGNomask, and hybrid models use specialized Triton kernels for structural attention. Table~\ref{tab:efficiency} reports throughput from the earlier H800 implementation, where structural models used FlexAttention~\cite{dong2024flex}. These measurements describe that implementation; throughput was not remeasured under a common setup for the RTX 5090 runs. Earlier campaign summaries report 133 hours on four H800 GPUs for the 100M variants, 125 hours on four H800 GPUs for the historical 500M campaign, and 385 hours on 64 A800 GPUs for the 1B variants.

\begin{table}[t]
  \centering
  \small
  \setlength{\tabcolsep}{4pt}
  \begin{tabular}{lrrr}
    \toprule
    Scale & Causal & Structural & Ratio \\
    \midrule
    100M & 62,175 & 37,975 & 1.6$\times$ \\
    500M & 12,550 & 8,316 & 1.5$\times$ \\
    \bottomrule
  \end{tabular}
  \caption{Historical H800 training throughput (token/s), using FlashAttention-2 for causal and FlexAttention for structural masks. The historical 100M mixing-head measurement was 3,375 token/s (approximately 18$\times$ slower than causal). These measurements do not characterize the new RTX 5090 runs.}
  \label{tab:efficiency}
\end{table}

\subsection{Tokenizer and Structural Tokens}
We use tokenizers with reserved non-terminal tokens (e.g., \texttt{NP)}, \texttt{(S}, and \texttt{VP)}).
The 100M and 500M models use the GPT-2 tokenizer~\cite{radford2019language}, and the 1B models use the Qwen3 tokenizer~\cite{yang2025qwen3}.
For tree linearization models, opening non-terminals (ONT) and closing non-terminals (CNT) are represented as reserved tokens so that linearized constituency structure is preserved during pretraining and evaluation.

\section{Evaluation Setup Details}
\label{sec:eval-details}

The BBC 100M and 500M models share the downstream and syntactic evaluation procedures below, including the same document-PPL test set. Model-specific exceptions are described with the corresponding task.

\subsection{Downstream Fine-Tuning on BBC News: XSum and BoolQ}
On BBC News, we evaluate XSum and BoolQ through task-specific fine-tuning from each pretrained checkpoint, with the optimizer state reset. We use the full XSum test set (11,333 examples) and BoolQ validation set (3,270 questions).
All models are fine-tuned with AdamW ($\beta_1{=}0.9$, $\beta_2{=}0.95$, weight decay 0.1) and a cosine learning rate schedule with 100 warmup steps decaying to $10^{-6}$.

\paragraph{XSum.} We fine-tune all models for 3 epochs with a peak learning rate of $6\times10^{-5}$ and a global batch size of 40.
During evaluation, we generate summaries with beam search using a beam size of 6 and a maximum length of 150 tokens.
For tree linearization models, beam search proceeds at the word level (word-sync) with a maximum of 75 word steps and a grammar constraint of at most 4 consecutive opening non-terminals.
ROUGE-1, ROUGE-2, and ROUGE-L are computed with the standard evaluate library; R-AVG denotes their arithmetic mean.

\paragraph{BoolQ.} We fine-tune all models for 5 epochs with peak LR $3\times10^{-4}$, global batch size 40.
We measure accuracy on the validation split.

The XSum and BoolQ entries in Tables~\ref{tab:exp1-results} and~\ref{tab:historical-bbc} are means $\pm$ sample standard deviations over five fine-tuning seeds (42, 2026, 6198, 13171, and 31723). Pushdown uses right-binarized, unary-collapsed silver spans. Its XSum decoder conditions on the source parse and jointly searches summary tokens and attachment actions with beam size 6 and no limit on consecutive reductions. BoolQ uses teacher-forced candidate scoring with supplied parser spans.

\subsection{Syntactic Generalization (SG)}
\label{sec:sg-eval-setup}
Following \citet{hu-etal-2020-systematic}, we evaluate the six SG test suites without fine-tuning by comparing target-region surprisal~\cite{hale-2001-probabilistic}.
For a tagged span $\mathbf{w}=\mathbf{x}_{s:e}$ and its left context $\mathbf{C}=\mathbf{x}_{<s}$, we define
\[
\begin{aligned}
S(\mathbf{w}\mid\mathbf{C})
  &= -\log p(\mathbf{w}\mid\mathbf{C}) \\
  &= -\log p(\mathbf{x}_{\leq e})+\log p(\mathbf{x}_{<s}).
\end{aligned}
\]
The logarithm base does not affect the SG predictions because the task-specific formulas only compare surprisals.

For terminal and pause-token models, we compute this span surprisal by summing token-level cross-entropy losses over the tagged span in one forward pass.
For tree-linearization models other than Tree-Shuffle, we approximate prefix probabilities with DFS word-synchronous beam search~\cite{stern-etal-2017-effective}.
At each synchronized terminal boundary $t$, the beam $\mathcal{B}_t$ contains the $b$ highest-scoring latent structural prefixes $\mathbf{y}^{(h)}_t$.
We form the beam-truncated marginal and the corresponding span score as
\[
\begin{aligned}
\widetilde{p}(\mathbf{x}_{\leq t})
  &= \sum_{h\in\mathcal{B}_t}
     p(\mathbf{x}_{\leq t},\mathbf{y}^{(h)}_t), \\
\widetilde{S}(\mathbf{w}\mid\mathbf{C})
  &= -\log\widetilde{p}(\mathbf{x}_{\leq e}) \\
  &\quad +\log\widetilde{p}(\mathbf{x}_{<s}).
\end{aligned}
\]
These span surprisals are the scores supplied to the task-specific formulas of \citet{hu-etal-2020-systematic}.

At each terminal step, the decoder must emit the observed next terminal, but may first expand non-terminal hypotheses through depth-first search.
Hypotheses synchronize after emitting the terminal, and the $b$ highest-scoring hypotheses are retained.
Following \citet{murty-etal-2023-pushdown} and \citet{sartran-etal-2022-transformer}, we set $b{=}300$; the within-step sub-beam is 300 and the terminal fast-track size is 30.
All such SG evaluations cap the linearized decoding length at $\max(6L,10)$, where $L$ is the terminal-token sequence length.
No separate limit is imposed on the total or consecutive number of non-terminal tokens.

Tree-Shuffle uses terminal-format teacher-forced scoring at test time. Pushdown instead incrementally marginalizes over attachment decisions with a beam size of 300.

\subsection{BLiMP}
BLiMP~\cite{warstadt-etal-2020-blimp-benchmark} evaluates whether models assign higher likelihood to grammatical sentences in minimal pairs across 12 grammatical phenomena.
For tree-linearization models, we obtain 300 parse proposals ($K=300$) from the Benepar parser~\cite{kitaev-klein-2018-constituency} and score each sentence by summing the joint probabilities of its candidate sequences. Tree-NoONT and Tree-Compress BLiMP scores have not been reevaluated with sequence deduplication; their reported values are marked accordingly.
A pair is scored correct if the good sentence receives higher marginalized probability than the bad sentence.
Each of the 67 subtasks contains 1,000 pairs.
For terminal and pause-token models, $K=1$ (no tree marginalization).
Tree-Shuffle likewise uses terminal-format scoring with $K=1$. For Pushdown, we convert the same proposals into right-binarized, unary-collapsed attachment sequences, retaining the intra-word BPE structure, and sum the joint token--attachment probabilities of distinct resulting candidates. Attachment probabilities are normalized over the sentence-local causal support used during training.

\subsection{Document-PPL Protocol}
\label{sec:ppl-protocol-details}

Both BBC scales use the same test set and document-PPL procedure. For each sentence, we sum the joint likelihoods of distinct valid linearized candidates without parser weights, following Eq.~\ref{eq:marginal}. The highest-probability candidate is retained as context for the next sentence, and histories reset at document boundaries. Perplexity is normalized by terminal and EOS tokens; BOS and padding positions are not scored.

Candidate sequences are deduplicated after each model's linearization, since NoONT and Compress can map different trees to the same sequence. Their corresponding BLiMP candidate checks remain pending, as marked in Table~\ref{tab:exp1-results}.

Tree-Shuffle uses the same PPL evaluation method as Terminal. Pushdown sums joint token--attachment probabilities and retains model-best histories; its architectural-reference results are separate from the controlled Tree/TG comparison. The supplementary BBC checkpoints use the same document-PPL protocol.

\subsection{OLMES Multiple-Choice Evaluation on FineWeb-Edu}
\label{sec:olmes-eval}
All 11 downstream tasks use the completion/cloze formulation standardized in OLMES~\cite{gu-etal-2025-olmes}. This formulation evaluates each answer string separately as a natural-language completion instead of predicting an answer label. All evaluations are multi-shot. We use 5-shot prompts for HellaSwag, WinoGrande, MMLU, MMLU-Redux, and OpenBookQA, and 3-shot prompts for BoolQ, ARC-Easy, ARC-Challenge, PIQA, SocialIQA, and CommonsenseQA.

We preprocess the natural-language content of every demonstration, target prompt, and answer option with Benepar~\cite{kitaev-klein-2018-constituency}.
For each parsed component $x$, we retain only its 1-best constituency parse $\hat{\mathbf{y}}(x)$ as a silver tree.
Depending on the task template, fields are parsed separately or the completed candidate sequence is parsed and split at the cloze boundary.
Task-specific template text is represented with the same constituency-tree notation.
The resulting trees are converted to the linearization required by each model.
Thus, the syntactic inputs differ in representation but share the same underlying terminal prompt and answer text.

For each demonstration, the parsed gold answer suffix is appended to its parsed prompt.
These demonstrations precede the target instance in every candidate prefix $\mathbf{p}_j$.
We then append the candidate-specific parsed suffix $\mathbf{c}_j$ and score only that suffix conditional on $\mathbf{p}_j$.
The prediction is $\arg\max_j S(\mathbf{c}_j \mid \mathbf{p}_j)$, using the task-specific option-score normalization. The primary results use $S_{\text{term}}$ in Eq.~\ref{eq:decom}; Appendix~\ref{sec:decom} also reports $S_{\text{full}}$.
For tasks using length normalization, we divide $S_{\text{term}}$ by the terminal continuation's character count; the remaining tasks use the raw sum.
For WinoGrande, the answer fills the blank in a candidate-specific prefix, while the common post-blank text is the scored suffix, following OLMES.
Unlike BLiMP, this MC protocol uses one silver tree per textual component and uses neither multiple parse proposals nor tree marginalization.

\section{Bootstrap Confidence Intervals and Paired Tests}
\label{sec:bootstrap}

We estimate uncertainty for the 1B FineWeb-Edu downstream evaluation using $B=10{,}000$ bootstrap resamples with seed 2027. For each task, examples are resampled with replacement and the same sampled indices are used for every model. Table~\ref{tab:bootstrap-ci} reports percentile 95\% confidence intervals for accuracy. The macro-average intervals instead resample tasks to measure sensitivity to task-suite composition. With shared task indices across models, TGTree ranks first in 52.18\% of resamples, Pause-2 in 42.27\%, and Pause-1 in 5.55\%. Excluding BoolQ changes these proportions to 27.00\%, 67.30\%, and 5.70\%, respectively. Terminal and Tree never rank first in either comparison.

\begin{table*}[!t]
  \centering
  \footnotesize
  \renewcommand{\arraystretch}{1.05}
  \setlength{\tabcolsep}{2.8pt}
  \begin{tabular}{lccccc}
    \toprule
    Task & Terminal & Tree & TGTree & Pause-1 & Pause-2 \\
    \midrule
    WG   & 58.17 [55.49, 60.93] & 61.64 [58.96, 64.33] & 63.38 [60.69, 66.06] & 58.96 [56.20, 61.64] & 62.59 [59.91, 65.27] \\
    HS   & 61.53 [60.57, 62.48] & 54.10 [53.14, 55.09] & 63.51 [62.57, 64.45] & 63.05 [62.09, 63.99] & 64.45 [63.51, 65.39] \\
    BQ   & 58.65 [56.97, 60.34] & 63.85 [62.23, 65.50] & 67.55 [65.93, 69.11] & 65.41 [63.82, 67.06] & 63.39 [61.71, 65.08] \\
    MM   & 36.79 [35.99, 37.59] & 36.88 [36.06, 37.69] & 37.76 [36.96, 38.56] & 37.70 [36.88, 38.51] & 38.38 [37.57, 39.18] \\
    MMR  & 37.61 [36.37, 38.89] & 37.53 [36.28, 38.77] & 38.42 [37.18, 39.70] & 38.81 [37.56, 40.07] & 39.46 [38.23, 40.72] \\
    OBQA & 43.60 [39.20, 48.00] & 44.00 [39.60, 48.40] & 45.80 [41.40, 50.20] & 42.80 [38.40, 47.00] & 44.00 [39.60, 48.40] \\
    CSQA & 53.24 [50.45, 56.02] & 56.02 [53.32, 58.80] & 58.15 [55.36, 60.93] & 56.18 [53.40, 58.97] & 56.76 [53.97, 59.54] \\
    SIQA & 48.98 [46.83, 51.23] & 48.46 [46.26, 50.67] & 50.31 [48.16, 52.56] & 51.28 [49.03, 53.58] & 48.67 [46.47, 50.92] \\
    ARCE & 69.30 [65.44, 72.98] & 68.42 [64.56, 72.11] & 70.00 [66.14, 73.68] & 74.04 [70.35, 77.54] & 72.98 [69.30, 76.49] \\
    ARCC & 43.48 [38.13, 49.16] & 45.15 [39.46, 50.84] & 44.82 [39.13, 50.50] & 43.48 [38.13, 49.16] & 48.16 [42.47, 53.85] \\
    PIQA & 73.88 [71.87, 75.90] & 73.07 [71.06, 75.08] & 74.97 [72.96, 76.93] & 75.14 [73.18, 77.09] & 74.97 [73.01, 76.93] \\
    \midrule
    AVG$^\dagger$ & 53.20 [46.38, 60.30] & 53.56 [46.79, 60.42] & 55.88 [48.39, 63.08] & 55.17 [47.86, 62.66] & 55.80 [48.68, 63.15] \\
    AVG w/o BQ$^\dagger$ & 52.66 [45.29, 60.26] & 52.53 [45.38, 60.00] & 54.71 [47.18, 62.60] & 54.14 [46.27, 62.17] & 55.04 [47.46, 62.91] \\
    \bottomrule
  \end{tabular}
  \caption{Accuracy point estimates and 95\% bootstrap confidence intervals for the terminal-format 1B evaluation. Per-task rows use paired resampling of evaluation examples. $^\dagger$Macro-average intervals resample tasks and therefore measure task-suite sensitivity rather than uncertainty within a fixed suite. Task abbreviations follow Table~\ref{tab:main-result}.}
  \label{tab:bootstrap-ci}
\end{table*}

Table~\ref{tab:bootstrap-tests} reports the paired contrasts most relevant to our claims: TGTree versus the Terminal baseline and each tree model versus its approximately budget-matched pause-token baseline. We compute two-sided bootstrap $p$-values from the paired difference distribution. The tests are exploratory and uncorrected for multiple comparisons; $p<.05$ is used only as a descriptive threshold.

\begin{table*}[!t]
  \centering
  \footnotesize
  \renewcommand{\arraystretch}{1.05}
  \setlength{\tabcolsep}{6pt}
  \begin{tabular}{lccc}
    \toprule
    Task & TGTree $-$ Terminal & Tree $-$ Pause-1 & TGTree $-$ Pause-2 \\
    \midrule
    WG   & +5.21 [2.29, 8.13]; $p<.001$ & +2.68 [$-$0.32, 5.68]; $p=.082$ & +0.79 [$-$2.05, 3.63]; $p=.604$ \\
    HS   & +1.98 [1.22, 2.74]; $p<.001$ & $-$8.94 [$-$9.96, $-$7.89]; $p<.001$ & $-$0.94 [$-$1.68, $-$0.19]; $p=.014$ \\
    BQ   & +8.90 [6.48, 11.31]; $p<.001$ & $-$1.56 [$-$3.27, 0.12]; $p=.074$ & +4.16 [2.02, 6.30]; $p<.001$ \\
    MM   & +0.97 [0.26, 1.67]; $p=.009$ & $-$0.82 [$-$1.53, $-$0.09]; $p=.027$ & $-$0.62 [$-$1.30, 0.08]; $p=.078$ \\
    MMR  & +0.81 [$-$0.32, 1.96]; $p=.172$ & $-$1.28 [$-$2.40, $-$0.18]; $p=.024$ & $-$1.04 [$-$2.14, 0.05]; $p=.069$ \\
    OBQA & +2.20 [$-$1.40, 5.80]; $p=.262$ & +1.20 [$-$2.80, 5.20]; $p=.592$ & +1.80 [$-$1.80, 5.40]; $p=.367$ \\
    CSQA & +4.91 [2.46, 7.45]; $p<.001$ & $-$0.16 [$-$2.46, 2.21]; $p=.934$ & +1.39 [$-$0.90, 3.69]; $p=.248$ \\
    SIQA & +1.33 [$-$0.67, 3.38]; $p=.210$ & $-$2.81 [$-$4.86, $-$0.77]; $p=.007$ & +1.64 [$-$0.36, 3.68]; $p=.120$ \\
    ARCE & +0.70 [$-$3.16, 4.56]; $p=.761$ & $-$5.61 [$-$9.30, $-$1.93]; $p=.002$ & $-$2.98 [$-$6.67, 0.70]; $p=.113$ \\
    ARCC & +1.34 [$-$4.01, 6.69]; $p=.666$ & +1.67 [$-$3.68, 6.69]; $p=.570$ & $-$3.34 [$-$9.03, 2.34]; $p=.261$ \\
    PIQA & +1.09 [$-$0.65, 2.83]; $p=.247$ & $-$2.07 [$-$3.92, $-$0.27]; $p=.024$ & +0.00 [$-$1.74, 1.69]; $p=1.000$ \\
    \bottomrule
  \end{tabular}
  \caption{Paired bootstrap differences in percentage points with 95\% confidence intervals and uncorrected two-sided $p$-values. Positive values favor the model named first. Tree--Pause-1 and TGTree--Pause-2 are approximate budget comparisons.}
  \label{tab:bootstrap-tests}
\end{table*}

\section{Historical BBC Experiments and Clean Reevaluation}
\label{sec:historical-bbc}

The two BBC scales in Table~\ref{tab:exp1-results} use different training datasets. \textbf{The 100M models were trained on the historical BBC stream; the ten 500M models were trained on the deduplicated BBC corpus} constructed in Appendix~\ref{sec:training}. The supplementary 500M models in Table~\ref{tab:historical-bbc} are separate historical runs. Historical BBC training comprises 15,187,326 article instances, including repeated and near-duplicate versions, with 10,061,025,584 terminal tokens of cumulative exposure (10.061B). The $\textsc{Lin}^1$ and $\textsc{Lin}^2$ budgets are approximately 24.71B and 32.03B input tokens. These training budgets differ from the corpus size in Table~\ref{tab:token-budgets}. TGNomask-Aug exposes CNT$_2$ representations to later tokens, unlike TGNomask.

A full token audit finds 2,071,964 distinct historical document sequences totaling 1,188,938,040 tokens (1.189B), each occurring 1--64 times. The 0.801B reference additionally merges normalized and near-duplicate versions and is separately encoded; it does not equal the exact unique content of the historical array. Both settings were trained for one epoch over their respective streams. Variable multiplicity does not correspond to uniform epochs over identical deduplicated examples.

An exact token comparison found complete training copies of 4,752 of the 4,966 original PPL test documents (95.69\%). We replace those perplexities using the test set described in Appendix~\ref{sec:ppl-protocol-details}, retaining the original XSum, BoolQ, and SG measurements. The 1,000-document development and 5,000-document test splits were filtered against the historical corpus using exact, near-duplicate, and local-reuse checks. The development split is used for validation during BBC 500M pretraining. A separate full scan of all 1,466,132 deduplicated training documents finds no complete token-identical matches to these development or test documents. The NoONT training arrays have the same terminal projection as this scanned corpus. This exact-token result does not rule out normalized-text matches, near duplicates, or shared fragments. The deduplicated corpus was separately parsed and encoded, so it is not assumed to be a byte-identical subset of the historical data.

\begin{table*}[!t]
  \centering
  \small
  \setlength{\tabcolsep}{7pt}
  \begin{tabular}{lccccc}
    \toprule
    Historical model & XSum$\uparrow$ & BoolQ$\uparrow$ & SG$\uparrow$ & BLiMP$\uparrow$ & Doc-PPL$\downarrow$ \\
    \midrule
    Terminal-500M & 20.75 $\pm$ 0.06 & 69.69 $\pm$ 0.60 & \textbf{71.11} & 64.74 & 17.303 \\
    Tree-500M & \textbf{22.25 $\pm$ 0.05} & 70.72 $\pm$ 0.35 & 63.13 & 76.09 & 16.001 \\
    TGTree-500M & 22.01 $\pm$ 0.08 & 70.63 $\pm$ 0.32 & 65.95 & \textbf{77.65} & 15.895 \\
    TGNomask-Aug-500M & 21.77 $\pm$ 0.06 & \textbf{70.78 $\pm$ 0.74} & 57.58 & 75.97 & \textbf{15.484} \\
    \bottomrule
  \end{tabular}
  \caption{Supplementary 500M models trained on the original BBC corpus. Evaluation follows Appendix~\ref{sec:eval-details}. XSum and BoolQ show means $\pm$ sample SD over five fine-tuning seeds. Best values are in \textbf{bold}.}
  \label{tab:historical-bbc}
\end{table*}

The historical Tree checkpoint has worse SG and document PPL at 500M than at 100M (63.13 vs.\ 81.40 and 16.001 vs.\ 14.736). The 500M Tree in Table~\ref{tab:exp1-results} reaches 83.43 SG and 13.864 PPL. Training corpora, token budgets, and optimization settings differ across these BBC runs and the FineWeb-Edu experiment; cross-setting comparisons therefore do not isolate the effects of model scale or data deduplication.

\section{SG Subtask Breakdown}
\label{sec:sg-breakdown}

Table~\ref{tab:sg-subtasks} breaks down SG performance for the ten BBC 500M models. Tree and TGTree both improve Agreement by 10.53 points over Terminal. Their differences emerge elsewhere: TGTree performs better on Garden Path Effects, while Tree performs better on Center Embedding and Long-Distance Dependencies. Tree-NoONT leads on Garden Path Effects and Licensing, which contribute to its highest overall SG score. TG and TGTree-Mix-TG have the highest Agreement scores but perform less well on Gross Syntactic Expectation and Long-Distance Dependencies.

\begin{table*}[!t]
  \centering
  \small
  \renewcommand{\arraystretch}{1.08}
  \setlength{\tabcolsep}{5pt}
  \begin{tabular}{lccccccc}
    \toprule
    Model & Agr$\uparrow$ & CtrEmb$\uparrow$ & GrdnPath$\uparrow$ & GrsSyn$\uparrow$ & Lic$\uparrow$ & LDD$\uparrow$ & SG$\uparrow$ \\
    \midrule
    Terminal & 64.91 & 91.07 & 78.29 & 94.57 & 78.95 & 77.52 & 80.88 \\
    Tree & 75.44 & 92.86 & 80.26 & 91.30 & 80.45 & \textbf{80.28} & 83.43 \\
    TGTree & 75.44 & 85.71 & 85.53 & 85.87 & 78.57 & 72.94 & 80.68 \\
    Pause-1 & 66.67 & 82.14 & 80.92 & \textbf{96.74} & 73.68 & 73.85 & 79.00 \\
    Pause-2 & 64.91 & 91.07 & 79.61 & 85.87 & 73.31 & 74.31 & 78.18 \\
    Tree-NoONT & 68.42 & 91.07 & \textbf{88.16} & 92.39 & \textbf{87.22} & 74.77 & \textbf{83.67} \\
    TG & \textbf{84.21} & 91.07 & 78.95 & 73.91 & 78.95 & 70.18 & 79.55 \\
    TGNomask & 70.18 & \textbf{94.64} & 82.89 & 79.35 & 83.08 & 79.36 & 81.58 \\
    TGTree-Mix-TG & \textbf{84.21} & 89.29 & 81.58 & 66.30 & 86.09 & 71.10 & 79.76 \\
    TGNomask-Mix-TG & 78.95 & 91.07 & 77.63 & 75.00 & 82.33 & 75.69 & 80.11 \\
    \bottomrule
  \end{tabular}
  \caption{SG subtask breakdown at 500M on BBC News. Agr = Agreement, CtrEmb = Center Embedding, GrdnPath = Garden Path Effects, GrsSyn = Gross Syntactic Expectation, Lic = Licensing, LDD = Long-Distance Dependencies. Best per metric in \textbf{bold}.}
  \label{tab:sg-subtasks}
\end{table*}

Table~\ref{tab:sg-subtasks-1b} reports the corresponding breakdown for the 1B FineWeb-Edu models. Tree and TGTree use the maximum linearized decoding length of six times the terminal sequence length described in Appendix~\ref{sec:sg-eval-setup}.

\begin{table*}[!t]
  \centering
  \fontsize{8pt}{10pt}\selectfont
  \setlength{\tabcolsep}{4pt}
  \begin{tabular}{lccccccc}
    \toprule
    Model & Agr$\uparrow$ & CtrEmb$\uparrow$ & GrdnPath$\uparrow$ & GrsSyn$\uparrow$ & Lic$\uparrow$ & LDD$\uparrow$ & SG$\uparrow$ \\
    \midrule
    Terminal & \textbf{87.72} & 94.74 & 93.42 & 98.91 & 74.81 & 78.44 & 88.01 \\
    Tree & 85.96 & \textbf{100.00} & \textbf{98.03} & \textbf{100.00} & 78.20 & 77.52 & \textbf{89.95} \\
    TGTree & 80.70 & \textbf{100.00} & 94.08 & \textbf{100.00} & \textbf{78.95} & \textbf{80.28} & 89.00 \\
    Pause-1 & 75.44 & \textbf{100.00} & 94.08 & 98.91 & 73.31 & 77.52 & 86.54 \\
    Pause-2 & 82.46 & 94.74 & 95.39 & \textbf{100.00} & 72.18 & 74.77 & 86.59 \\
    \bottomrule
  \end{tabular}
  \caption{SG subtask breakdown at 1B on FineWeb-Edu. Best per metric in \textbf{bold}; abbreviations follow Table~\ref{tab:sg-subtasks}.}
  \label{tab:sg-subtasks-1b}
\end{table*}

\section{Full-versus-Terminal Score Sensitivity}
\label{sec:decom}

We compare the raw full and terminal scores in Eq.~\ref{eq:decom} on an eight-task suite (Validation-8; 17,691 examples): WinoGrande, HellaSwag, OpenBookQA, CommonsenseQA, SocialIQA, ARC-Easy, ARC-Challenge, and PIQA. Except for OpenBookQA validation, these are the same splits used in the main evaluation. Fixed in-context demonstrations come from training data and do not overlap with the scored examples. Table~\ref{tab:decom} reports predictions based on raw log-probability sums for both models, without the task-specific length normalization used in Table~\ref{tab:main-result}.

\begin{table*}[!t]
  \centering
  \small
  \renewcommand{\arraystretch}{1.08}
  \setlength{\tabcolsep}{5pt}
  \begin{tabular}{lrrr@{\hspace{12pt}}rrr}
    \toprule
    & \multicolumn{3}{c}{Tree} & \multicolumn{3}{c}{TGTree} \\
    \cmidrule(lr){2-4}\cmidrule(lr){5-7}
    Task & Full & Term & $\Delta$ & Full & Term & $\Delta$ \\
    \midrule
    WinoGrande & 61.17 & 61.64 & +0.47 & 64.64 & 63.38 & $-$1.26 \\
    HellaSwag & 37.83 & 43.92 & +6.08 & 47.11 & 49.45 & +2.34 \\
    OpenBookQA & 33.60 & 33.60 & +0.00 & 34.20 & 34.00 & $-$0.20 \\
    CommonsenseQA & 54.79 & 55.86 & +1.06 & 57.17 & 57.49 & +0.33 \\
    SocialIQA & 43.65 & 45.60 & +1.94 & 44.42 & 45.80 & +1.38 \\
    ARC-Easy & 71.05 & 74.74 & +3.68 & 73.68 & 74.39 & +0.70 \\
    ARC-Challenge & 39.80 & 41.81 & +2.01 & 38.80 & 39.80 & +1.00 \\
    PIQA & 70.24 & 72.25 & +2.01 & 72.63 & 73.29 & +0.65 \\
    \midrule
    Macro average & 51.52 & 53.68 & +2.16 & 54.08 & 54.70 & +0.62 \\
    \bottomrule
  \end{tabular}
  \caption{Accuracy (\%) using raw full and terminal scores on the eight-task sensitivity suite; $\Delta=\mathrm{Term}-\mathrm{Full}$. Macro averages are unweighted. Paired 95\% bootstrap intervals for the macro $\Delta$ are [1.35, 2.99] for Tree and [$-$0.09, 1.35] for TGTree (10,000 resamples; seed 2026).}
  \label{tab:decom}
\end{table*}

Omitting structural-token scores raises the macro average by 2.16 points for Tree and 0.62 points for TGTree; the latter interval includes zero. For Tree, 7.46\% of predictions change from wrong to correct and 5.30\% change in the opposite direction. The corresponding TGTree rates are 4.92\% and 4.30\%. The benefit varies by task, with two TGTree tasks favoring full scoring. These results characterize sensitivity to structural-token likelihoods under raw scoring. Because the analysis shares evaluation examples and uses different normalization, it does not independently validate the main scoring rule.

The two scores address related but distinct questions. $S_{\text{full}}$ scores the candidate together with one parser-selected silver tree. $S_{\text{term}}$ accumulates only terminal-token factors while retaining that same tree as context. Neither is the exact marginal sentence likelihood $\sum_{\mathbf{y}}p(\mathbf{x},\mathbf{y})$.

\subsection{Alternative Ten-Task Comparison}
\label{sec:validation10}

Validation-10 adds BoolQ and MMLU validation to the eight tasks above, yielding 22,207 scored examples, and applies the task-specific terminal-score normalization. Eight tasks share 20,461 examples with the main evaluation (92.1\% of this suite); OpenBookQA and MMLU use different splits. This is a sensitivity comparison rather than an independent evaluation set. For MMLU, five fixed demonstrations per subject are removed from scoring, leaving 1,246 examples across 57 subjects. MMLU accuracy is macro-averaged across subjects; other task accuracies are micro-averaged. The headline score averages the ten tasks, with 10,000 paired bootstrap resamples and seed 2026.

\begin{table*}[!t]
  \centering
  \small
  \renewcommand{\arraystretch}{1.08}
  \setlength{\tabcolsep}{5pt}
  \begin{tabular}{lcc}
    \toprule
    Model & Macro accuracy & $\Delta$ vs.\ Terminal (95\% CI) \\
    \midrule
    Terminal & 54.43 & reference \\
    Tree & 54.75 & +0.32 [$-$0.67, 1.30] \\
    TGTree & \textbf{57.35} & +2.92 [1.95, 3.84] \\
    Pause-1 & 56.47 & +2.05 [1.15, 2.93] \\
    Pause-2 & 57.09 & +2.66 [1.82, 3.49] \\
    \bottomrule
  \end{tabular}
  \caption{Terminal-score accuracy (\%) on Validation-10. The paired TGTree--Pause-2 difference is +0.26 points (95\% CI [$-$0.68, 1.20]).}
  \label{tab:validation10}
\end{table*}

% Keep the final appendix tables together at the top of their float page.
\begingroup
\makeatletter
\setlength{\@dblfptop}{0pt}
\setlength{\@dblfpsep}{18pt}
\makeatother
\clearpage
\endgroup

\end{document}